\documentclass{llncs}

\usepackage[utf8]{inputenc}
\usepackage{graphicx}
\usepackage{url}
\usepackage{subfigure}
\usepackage{multirow}
\usepackage{array}
\usepackage{amsmath}

\newcommand*\samethanks[1][\value{footnote}]{\footnotemark[#1]}

\title{Analyzing First-Person Stories Based on Socializing, Eating and Sedentary Patterns}
\author{Pedro Herruzo\thanks{These authors contributed equally to this work.} \and Laura Portell\samethanks \and Alberto Soto\samethanks \and Beatriz Remeseiro}

\institute{Departament de Matem\`atiques i Inform\`atica, Universitat de Barcelona\\ Gran Via de les Corts Catalanes 585, 08007 Barcelona, Spain\\
\email{pedro.herruzo@ub.edu, portell.laura@gmail.com, alsoba13@gmail.com, bremeseiro@ub.edu}}

\begin{document}
\maketitle

\begin{abstract}
First-person stories can be analyzed by means of egocentric pictures acquired throughout the whole active day with wearable cameras. This manuscript presents an egocentric dataset with more than 45,000 pictures from four people in different environments such as working or studying. All the images were manually labeled to identify three patterns of interest regarding people's lifestyle: socializing, eating and sedentary. Additionally, two different approaches are proposed to classify egocentric images into one of the 12 target categories defined to characterize these three patterns. The approaches are based on machine learning and deep learning techniques, including traditional classifiers and state-of-art convolutional neural networks. The experimental results obtained when applying these methods to the egocentric dataset demonstrated their adequacy for the problem at hand.
\end{abstract}

\subsubsection{Keywords:} first-person stories, wearable cameras, egocentric lifelogging, annotation tool, deep learning, machine learning.

\section{Introduction}

Egocentric lifelogging is a recently new research field that consists in capturing daily experiences of people from continuous records taken by them \cite{bolanos2017toward}. Egocentric vision is the next step on the development of the lifelogging technology, since it provides additional visual information taken from wearable cameras using a first person point-of-view. Egocentric visual data analysis can generate useful information about a person on different areas such as social interaction \cite{alletto2014ego,aghaei2016whom}, food localization and recognition \cite{bolanos2016simultaneous}, sentimental analysis \cite{talavera2017sentiment}, etc. 

Three different kinds of groups are interested in egocentric data analysis. First audience corresponds to people that want to quantify their lifestyle, the so-called quantified self community. The second group are professionals, such as doctors that use this technology to observe active aging of older people or to create cognitive exercises for patients with Alzheimer's disease \cite{doherty2011automatically}. The last group is formed by influential people, such as elite athletes who wear head-mounted cameras like GoPro\footnote{https://gopro.com/} to remember their emotional experiences \cite{hoshen2016egocentric}. 

The egocentric vision field is an emerging field that has recently become increasingly active. There are several works that try to face different topics in this area of research. For the analysis of social interactions, Alletto et al. \cite{alletto2014ego} build a model that estimates head pose and 3D location in egocentric video sequences; and Aghaei et al. \cite{aghaei2016whom} exploit the distance and the orientation of the appearing individuals using pictures. Regarding activities of daily living, Cartas et al. \cite{cartas2017recognizing} explore their classification in 21 categories, that includes eating and socializing activities, using egocentric images and convolutional neural networks. 

The performance of any machine learning and/or computer vision method depends on the quality and quantity of the training data. However, there are not many proposed datasets for egocentric vision, specially, datasets with low temporal resolutions. Some examples of egocentric datasets include: GTEA \cite{fathi2011learning}, a dataset of videos acquired with a GoPro camera and captured by four different subjects, which contains seven types of daily activities and each video is labeled with the list of objects involved and background segmentations; EDUB-Seg \cite{dimiccoli2017sr}, a low-temporal resolution egocentric dataset acquired by the Narrative Clip camera, which includes 18,735 images captured by seven users during 20 days and includes indoor and outdoor scenes with numerous foreground and background objects manually annotated to provide a temporal segmentation ground-truth; and Egocentric Food \cite{bolanos2016simultaneous}, the first dataset of egocentric images for food-related objects localization and recognition that contains 5,038 images collected using the Narrative Clip camera, 8,573 bounding boxes and 9 different food classes.

In order to analyze people's lifestyle patterns during long periods, it is necessary to take pictures for at least 10 hours periods. Taking that into account, we present an egocentric dataset composed of more that 45,000 images taken from four people who wore the camera during active hours. In addition, we propose a research methodology to extract useful information about three different patterns: socializing, eating and sedentary. The proposed methodology should be able to quantify the following information: 1) social patterns, such as time spent with other people; 2) eating patterns, such as timing of meals and duration; and 3) sedentary lifestyle patterns, such us time spent sitting at a desk. Furthermore, these three patterns can be combined allowing to determine information such as time spent eating alone or with other people. 

The remainder of the paper is organized as follows: Sect. \ref{sec:materials_methods} presents the egocentric dataset and the proposed methods for pattern classification, Sect. \ref{sec:results_discussion} presents the experimentation performed and the validation results, and finally, Sect. \ref{sec:conclusions} includes the conclusions and future lines of research.

\section{Materials and Methods} \label{sec:materials_methods}

In this section, first we present our egocentric dataset and the adapted annotation tool that allowed us to set the ground truth for each image. Second, we explain the different approaches that we used to achieve our objectives.

\subsection{Egocentric Dataset} \label{sec:dataset}

Due to the lack of first-person images to analyze socializing, eating and sedentary lifestyle patterns, we have created a dataset called LAP. It is made of egocentric pictures taken from a Narrative Clip\footnote{http://getnarrative.com/} camera and contains 45,297 images taken from four different people in consecutive days with a frame rate of 2 fpm. Each person took the pictures in very different contexts such as working, studying or vacation. All the images were manually labeled according to the three following patterns:

\begin{itemize}
\item Eating pattern, three labels: eating (E), food related non eating (FRNE), non food related (NFR). Whereas other works can only distinguish between food or not food, this dataset allows to discard false positives when there is an image containing food but the subject is not eating.
\item Socializing pattern, two labels: socializing (S), not socializing (NS). This problem was simplified as being with a person or not. As a limitation of this approach, we cannot distinguish the false positives when there are people around the subject who are not interacting with him/her.
\item Sedentary pattern, two labels: table (T), no table (NT). This problem was simplified by determining if he/she is in front of a table or not, which is strongly related to the sedentary pattern of being sat in front of a table. 
\end{itemize}

\begin{figure}[htb]
  \centering
  \includegraphics[width=1.\textwidth]{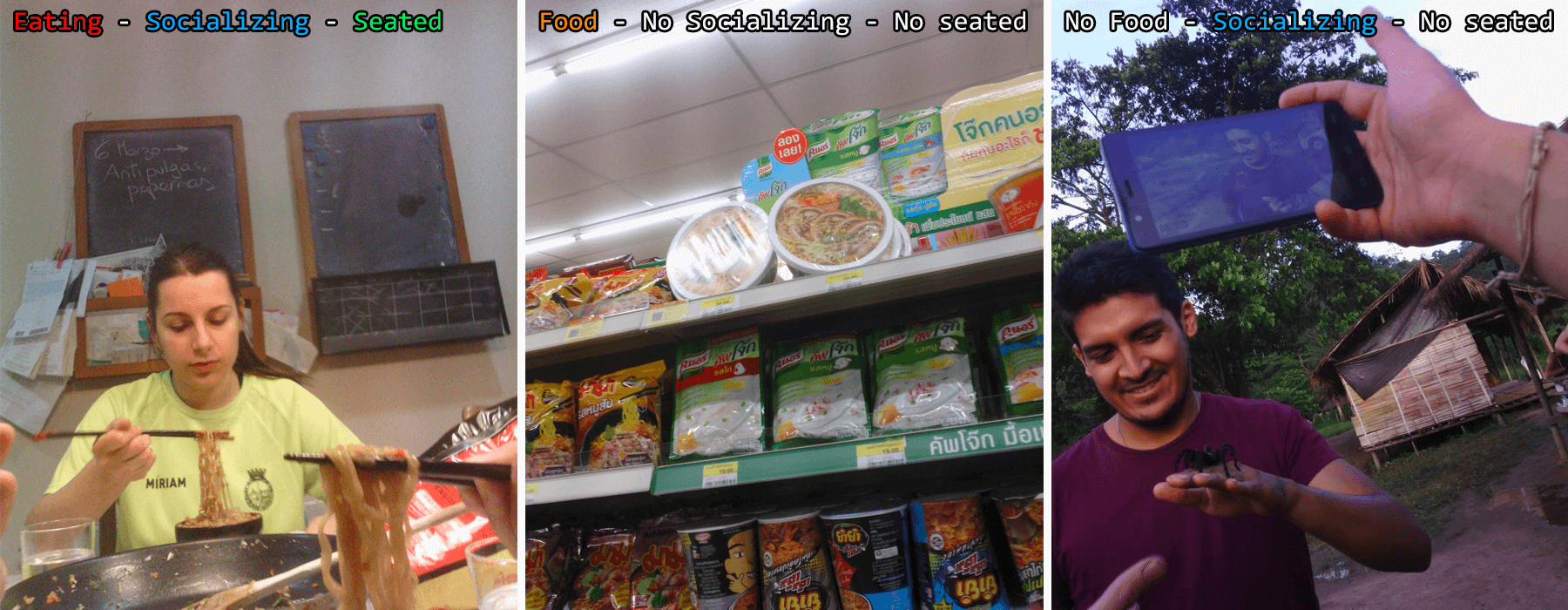}
  \caption{Example of three egocentric images of the LAP dataset, each one with different labels (see the top of each picture).}
  \label{fig:egocentricPicture}
\end{figure}

Each picture had to be assigned with three labels, one per each of the previous sets. As labeling all pictures requires a reasonable amount of time, we have built LAP annotation-tool, a specialized annotation tool with many keyboard shortcuts, which has been developed by adapting the web-based tool for image annotation known as LabelMe \cite{russell2008labelme,torralba2010labelme}. LAP annotation-tool allows to load pictures and select $N$ sets of labels, and then it creates an environment with $N$ keys to switch between the target labels. For the problem at had, three different sets of labels were needed ($N=3$) and so we used three keys (numbers $1$, $2$, and $3$) to switch between the different labels, setting always one label per set. Figure \ref{fig:egocentricPicture} shows three representative images of the LAP dataset with their respective assigned labels. Note that the LAP annotation-tool can be used for any image annotation problem with multiple labels per image, and it is available for download from our Github\footnote{https://github.com/alsoba13/LAP-Annotation-Tool}.

During the process of labeling, a set of rules for data integrity was established to avoid different labels in images that represent the same scene:

\begin{itemize}
\item Eating pattern: E is used when there is food in the image and the person is eating, FRNE is used when there is food in the image but the subject is not eating, and NFR is used when there is no food in the image.
\item Socializing pattern: S is only used when a person appears in the image, regardless of the distance.
\item Sedentary pattern: T is only used when a table appears in the image and it is not far from the camera wearer.
\end{itemize}

Regarding noisy or black pictures, instead of discarding them, they were assigned the default labels NFR-NS-NT. In this manner, the trained model should be able to consider this situation that frequently occurs in real environments.

Table \ref{tab:split} shows some statistics for the LAP dataset taking into account all the possible combinations among the three sets of labels. First insights of data show that the dataset is highly imbalanced. This fact was expected since, in real life, people do not spend the same amount of time socializing than alone, or eating than doing other daily routines or activities. If the different combinations of labels are analyzed, it can be observed that there are several combinations poorly represented. Note that only 4 out of the 12 combinations represent over the 92\% of the total number of images acquired.

\begin{table}[htb]
\centering
\caption{Distribution of classes in the LAP dataset.}
\label{tab:split}
\begin{tabular}{m{0.25in}m{1in}m{0.45in}m{0.55in}}
\hline
Id & Labels & \% & \#Images \\ \hline
0 & NFR-NS-NT & 46.49  & 21,058 \\
1 & NFR-NS-T & 12.97 & 5,877 \\ 
2  & NFR-S-NT & 21.53 & 9,755 \\ 
3 & NFR-S-T & 11.46 & 5,194 \\ 
4 & FRNE-NS-NT & 0.41 & 187 \\ 
5 & FRNE-NS-T & 0.48 & 218 \\
6 & FRNE-S-NT & 0.94 & 425 \\
7 & FRNE-S-T & 1.49 & 673 \\
8 & E-NS-NT & 0.19 & 88 \\
9 & E-NS-T & 1.20 & 543 \\
10 & E-S-NT & 0.53 & 242 \\ 
11 & E-S-T & 2.29 & 1,037 \\ \hline
& \textbf{Total} & \textbf{100} & \textbf{45,297} \\ \hline
\end{tabular}
\end{table}

For experimental purposes, the LAP dataset has been split in training, validation and test sets: the training set contains a 70\% of the images, whilst the validation and test sets contain, each one, a 15\%.

\subsection{Methods} \label{sec:methods}

Given an input image, the goal is to classify it in order to determine the three patterns of interest: socializing, eating and sedentary. Accordingly, the following sets were defined: $\textit{Eating}:=\{\textit{E},\textit{FRNE},\textit{NFR}\}$, $ \textit{Socializing}:=\{\textit{S},\textit{NS}\}$, and $\textit{Sedentary}:=\{\textit{T},\textit{NT}\}$. All the possible combinations of the three sets are considered, resulting in the cartesian product $\textit{Eating} \times  \textit{Socializing} \times \textit{Sedentary}$, a set with $3\times 2\times 2=12$ classes. Therefore, we have a 12-class classification problem for which two different approaches have been considered based on machine learning and deep learning techniques. The two approaches are subsequently presented.

\subsubsection{Machine Learning Approach.} 
\begin{figure}[htb!]
  \centering
  \includegraphics[width=\textwidth]{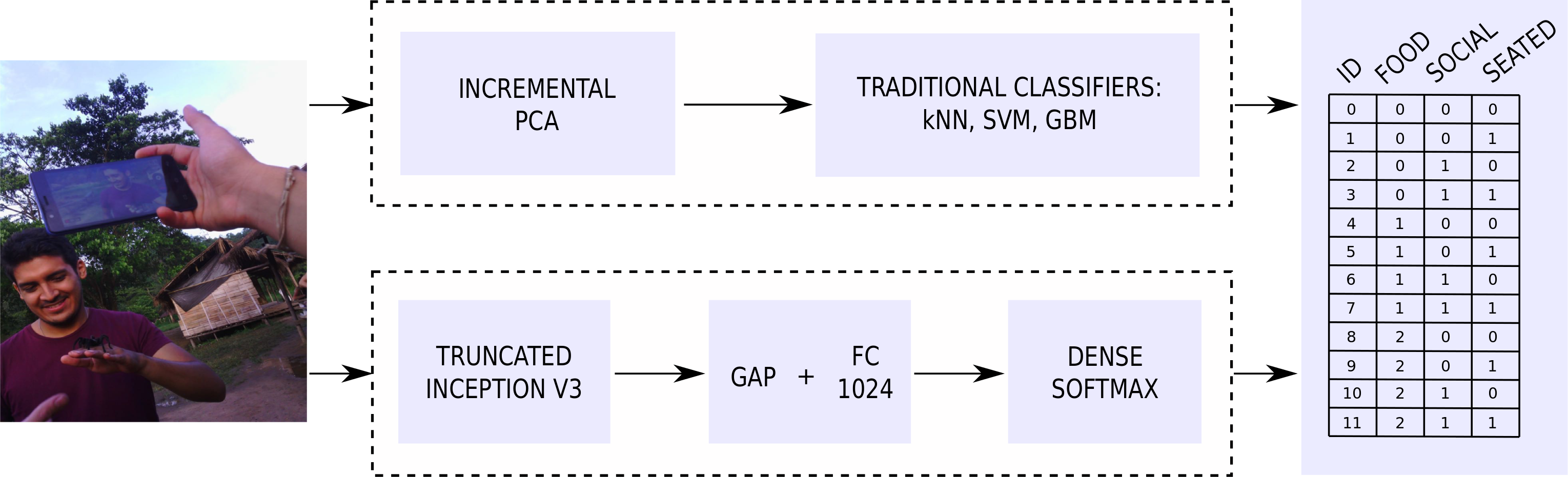}
  \caption{Workflow of the machine learning (top) and deep learning (bottom) approaches.}
  \label{fig:methods}
\end{figure}

The first approach is depicted in Fig. \ref{fig:methods} (top) and consists in using machine learning (ML) algorithms to classify an input image into one of the 12 target categories. 

Applying classical ML methods directly to images requires the use of a feature extraction step before the classification. For this task, we have used the incremental principal component analysis (IPCA) technique \cite{balsubramani2013fast}, which projects the data into a reduced space computing the projection matrix iteratively. Next, three well-known algorithms were used for classification, which were selected aiming to analyze different approaches of the supervised learning process:

\begin{itemize}
\item $k$-nearest neighbors ($k$NN) \cite{kramer2013k}: this method assigns the class label of the majority of the $k$ nearest patterns in the data space, based on the idea that the nearest patterns to a target one deliver useful information.
\item Support vector machines (SVM) \cite{burges1998tutorial}: they are based on the statistical learning theory and revolve around the notion of a \textit{margin}, either side of a hyperplane that separates two classes.
\item Gradient boosting machines (GBM) \cite{natekin2013gradient}: they are powerful techniques for both regression and classification problems, which can be seen as ensembles of weak prediction models, typically decision trees.
\end{itemize}

\subsubsection{Deep Learning Approach.} The second approach is illustrated in Fig. \ref{fig:methods} (bottom) and aims at classifying an input image using deep learning algorithms. 

Convolutional neural networks (CNNs) were considered in this case and, more specifically, the deep architecture known as InceptionV3 \cite{szegedy2016rethinking}. It is a general model for any kind of images with an only assumption about their size: the dimensions of the input layer are $299 \times 299 \times 3$, allowing to compute all convolutions with a valid size after the reductions made by pooling layers. 

This model was first pre-trained on a large dataset called ImageNet \cite{russakovsky2015imagenet}. Then, the last layers were adapted to our 12-class classification problem. Basically, the last fully connected, pooling and vectorizing layers were removed from the original model; whilst a global average pooling (GAP) and a 1024-unit fully connected (FC) layers were added before the last fully connected layer with a softmax. Additionally, batch normalization \cite{ioffe2015batch} and dropout \cite{srivastava2014dropout} were added to our deep learning approach to avoid the overfitting shortcoming of the CNNs.

The binary cross entropy \cite{lei2015predicting} was used in our model as the loss function. In order to fix the problem of imbalanced classes, described in Sect. \ref{sec:dataset}, we combined the use of weights with the loss function. The weights were defined as:  

\begin{equation}
w_i = \frac{M}{N_i}
\end{equation}
where $N_i$ is the number of images in class $i$ ($i\in \textit{Eating} \times \textit{Socializing} \times \textit{Sedentary}$), and $M$ is the number of pictures of the major class ($M = \underset{i}{max} N_i$).

\section{Experiments and Discussion} \label{sec:results_discussion}

This section includes the evaluation of our methods using the LAP dataset previously presented, in addition to some details about the experimental setup and the performance measures considered.

\subsection{Experimental setup}

Experimentation was carried out on a Intel\textsuperscript{\copyright} Core\textsuperscript{\texttrademark} i7-6700 CPU @ 8M Cache, 3.40 GHz with RAM 32 GB DDR4. For the deep learning approach, a NVIDIA TITAN Xp GPU was also used.

Regarding the machine learning approach, the Scikit-learn library \cite{pedregosa2011scikit} was used to train the three classifiers. Their configuration parameters were selected by merging the training and validation sets, and then applying grid search and 3-fold cross validation. The following configuration was finally used: $k$NN classifier with number of neighbors $k=3$, SVM with linear kernel and penalty parameter $C=100$, and GBM with regression trees as weak prediction models.

With respect to the framework for the deep learning approach, we used Keras \cite{chollet2015keras}, a Python deep learning library for Theano and TensorFlow. In particular, we run it on top of TensorFlow \cite{abadi2016tensorflow}. Model selection of the CNN approach was made by training the network over the training set and selecting the parameters that make a better score and less overfitting over the validation set. The architecture and training details are as follows: a stochastic gradient descent for optimization half of the epochs and Adam \cite{DBLP:journals/corr/KingmaB14} the rest, both with learning rate of 0.001, a momentum of 0.9, and a batch size of 128 images. Additionally, the CNN was trained over 50 epochs, and data augmentation was applied with flipping, Gaussian noise, and a rotation from $-30$ up to $30$ degrees. 

In order to match the model requirements of Inception V3, the images of our dataset were reduced from $1944 \times 2592 \times 3$ to $299 \times 299 \times 3$. Note that this reduction of the input images was applied in both approaches in order to get a fair comparison of the results.

\subsection{Performance measures}

Three different metrics were used to evaluate the adequacy of the proposed methods for the classification of the socializing, eating and sedentary patterns:

\begin{itemize}
\item F1-score: the harmonic mean of precision and recall.
\item Accuracy: the percentage of correctly classified samples.
\item Normalized accuracy: the weighted accuracy in which each class contributes with the ratio of correct predictions over the total of images, normalizing by the number of classes.
\end{itemize}

It should be pointed out the relevance of the normalized accuracy since the dataset is highly imbalanced, and so this metric allows us to know how good is the method classifying each class in a more precise way.

\subsection{Results}

Table \ref{tab:results} shows the classification results for the task of predicting the class of an input image. Note that the best results appear in bold face.

Regarding the machine learning approach, based on incremental PCA and traditional classifiers, $k$NN and SVM have a quite similar behavior in terms of performance with a F1-score over 0.3 and an accuracy over the 40\%. The best result obtained in this case corresponds to GBM, with a F1-score close to 0.5 and an accuracy over the 53\%. If the normalized accuracy provided by the three classifiers is analyzed, the results are quite poor due to the imbalance of the dataset (a normalized accuracy of 15.11\% in the best case). In particular, the images labeled as \textit{NFR-NS-NT} correspond to the 46.5\%, so it could be said that this class is mainly the only one learned by these models. Note that this behavior is also related with the low F1-score, due to the poor precision obtained when comparing the major class with the others. 

\begin{table}[htb]
\centering
\caption{Results for the machine learning (ML) and deep learning (DL) approaches.}
\label{tab:results}
\begin{tabular}{m{1.3in}m{0.5in}m{0.5in}m{0.5in}@{\extracolsep{7pt}}m{0.7in}m{0.7in}}
\cline{2-6}
& \multicolumn{3}{c}{ML approach} & \multicolumn{2}{c}{DL approach} \\ \cline{2-4} \cline{5-6} 
& $k$NN & SVM & GBM & non-weights & weights \\ \hline
F-1 score & 0.355 & 0.368 & 0.490  & 0.309 & \textbf{0.64}\\ 
Accuracy (\%) & 49.25 & 42.52 & 53.72& 46.75 & \textbf{60.53}\\ 
Normalized acc. (\%) &10.11 &9.69 &15.11 & 8.59 & \textbf{57.55}\\ \hline
\end{tabular}
\end{table}

With respect to deep learning, the results obtained without considering the weights are quite similar to the ones provided by the classical machine learning methods. As a matter of fact, the use of GBM as classifier in the ML approach outperforms the basic DL approach despite the fact that IPCA only does a space reduction on raw pixels data instead of getting more abstract representations. However, when using the proposed weights as part of the binary cross entropy loss function, in order to face the problem of imbalanced classes, these measures are noticeably improved. In particular, the F1-score obtained is 0.64 and the accuracy surpasses the 60\%. With respect to the normalized accuracy, it is almost aligned with the accuracy since it reaches the 57\%. This result should be highlighted since it is almost four times better than the maximum normalized accuracy obtained in the best configuration of the ML approach (15.11\%) and almost seven times better than the one obtained in the first DL approach (8.59\%), which demonstrated the key role played by our proposed weights.

\begin{figure}[htb!]
  \centering
  \begin{minipage}{.35\textwidth}
  \centering
  \includegraphics[width=\linewidth]{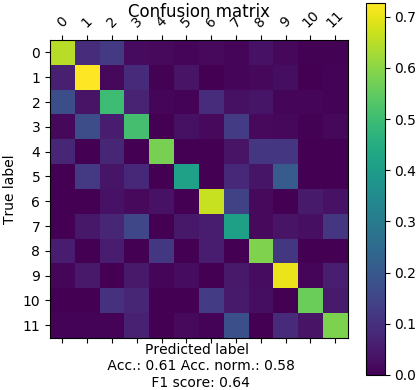}
  \caption{Confusion matrix of the deep learning approach using our proposed weights to classify input images into the 12 classes (see Table \ref{tab:split} for a detailed explanation of each \textit{id}).}
  \label{fig:confusion_matrix}
\end{minipage}
\hfill
\begin{minipage}{.59\textwidth}
  \centering
  \includegraphics[width=\linewidth]{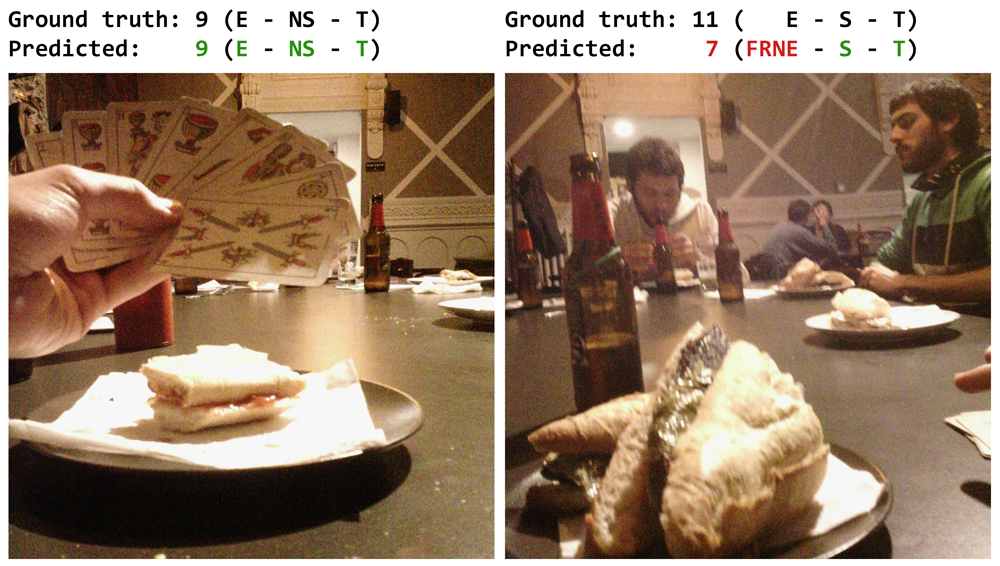}
  \caption{Two images of the LAP dataset with the ground truth and the predicted labels: (left) a correct classification of an image from class 9; and (right) a misclassification of an image labeled as \textit{S}, so the food in it may correspond to food being eaten by the subject (\textit{E}) or by other people (\textit{FRNE}).} \label{fig:errors}
\end{minipage}
\end{figure}

Figure \ref{fig:confusion_matrix} displays the confusion matrix of the deep learning approach when the weights are used as part of the binary cross entropy loss function. As can be seen, most of the error comes from misclassifications on the \textit{Eating} pattern. For example, class 11 (\textit{E-S-T}) is often classified as 7 (\textit{FRNE-S-T}), so the model just makes a mistake with the \textit{Eating} component of the triplet. This fact also happens with classes 5 (classified as 1 and 9), 6 (classified as 2 and 10), 7 (classified as 3 and 11) and 10 (classified as 6). All those errors form visible lower and upper diagonals in the confusion matrix. After a more careful analysis, it can be also observed that this trend has an exception since our model correctly classifies most of the samples from classes 8 (\textit{E-NS-T}) and 9 (\textit{E-NS-NT}). Images of these two classes have in common that contain food (\textit{E}) but no people (\textit{NS}). Therefore, it can be said that, when classifying an image of any of these two classes, the model has a strong belief on the person is eating (\textit{E}) as he/she is alone. On the other hand, food in \textit{S} images may be from the subject itself or from any of his/her companions, which makes that these images are sometimes misclassified as \textit{FRNE}. Figure \ref{fig:errors} shows two images from the LAP dataset with both the ground truth and the predicted labels.

\section{Conclusions} \label{sec:conclusions}

First-person cameras are inherently linked to the ongoing experiences of the people who wear them. Pictures acquired by this type of cameras allow to analyze the visual world with respect to the wearer's activities and behaviors.

In this context we present LAP, an egocentric dataset composed of 45,297 pictures taken from four subjects using a wearable camera. In addition to the first-person images, the dataset contains three labels per picture that correspond to the three patterns of interest: socializing, eating and sedentary. Furthermore, we present a simple, yet very useful annotation tool based on LabelMe that allows us to label pictures with more than one label in a very reasonable time. 

Regarding the research methodology, we have proved that we can estimate socializing, eating and sedentary patterns of a subject from egocentric pictures by combining different powerful methods and adapting them to our problem. More specifically, a preliminary comparison of two approaches was presented, one of them based on classical machine learning algorithms and the other one on state-of-art deep learning techniques. Both approaches were evaluated over the LAP dataset using three performance measures: F1-score, accuracy and normalized accuracy. The obtained results demonstrated the adequacy of the proposed methods to solve this multi-class problem. It should be highlighted that the deep learning approach outperforms the classical machine learning methods due to the complexity of the problem, with 12 classes and a highly-imbalanced dataset. In particular, the use of the proposed weights in conjunction with the binary cross entropy loss function allows us to achieve the most competitive results, with a normalized accuracy over 57\%.

As future work, we plan to explore a multi-task approach in order to predict the socializing, eating and sedentary patterns. On the other hand, the problem of estimating the sedentary lifestyle of a person, i.e. if he/she is sitting or walking, is very difficult to predict in short-term. For this reason, the future research also includes to introduce time dependency in our models. Finally, we would like to increase the labels of the LAP dataset by including new information such as the number of hours spent with a smartphone.

\section*{Acknowledgements}

This work was partially funded by TIN2015-66951-C2-1-R, SGR 1219, and CERCA Programme / Generalitat de Catalunya. Beatriz Remeseiro acknowledges the support of the Ministerio de Econom\'ia y Competitividad of the Spanish Government under \textit{Juan de la Cierva} Program (ref. FJCI-2014-21194). The funders had no role in the study design, data collection, analysis, and preparation of the manuscript.

The authors gratefully acknowledge the support of NVIDIA Corporation with the donation of the Titan Xp GPU used for this research.

\end{document}